\documentclass{article}
\usepackage{xcolor}
\usepackage{amssymb}
\usepackage{spconf,amsmath,graphicx,hyperref}

\title{Mechanistic Interpretability Reveals Shared Causal Subspaces in Brain-to-Speech Decoders}
\name{Maryam Maghsoudi$^{\star}$ \qquad Ayushi Mishra$^{\dagger}$ \qquad Sanghamitra Dutta$^{\star}$}
\address{$^{\star}$Department of Electrical and Computer Engineering, University of Maryland, College Park, MD, USA \\
	$^{\dagger}$Department of Computer Science, University of Maryland, College Park, MD, USA}
\begin{document}
%
\maketitle
\begin{abstract}
Decoding covert speech, such as mimed or imagined, from brain activity is harder than decoding vocalized speech. Cross-modal transfer, where information from one speech form helps decode another, is a promising remedy; yet how a decoder internally represents and processes brain activity from different speech forms remains unclear. In this work, we ask: which internal neurons of a decoder carry cross-modal information, and are these neurons shared across different speech forms? To answer these questions, we leverage mechanistic interpretability, using recordings of the same sentences in vocalized, mimed, and imagined input pairs for activation patching. We insert the decoder's internal activity for a sentence in one condition into its processing of the same sentence in another and measure the change in decoding accuracy. We find that no single neuron drives this benefit; instead, it arises from small groups of neurons, with vocalized speech as the most useful source. These groups are largely condition-specific in the early stage of the decoder but overlap in the later stage. These findings point toward more data-efficient covert speech decoders through training objectives that encourage shared later-stage representations learned mainly from vocalized data.
\end{abstract}


\section{Introduction}
\label{sec:intro}
Neural networks can now reconstruct speech from brain activity recorded during both overt and covert speech, such as silently articulating (mimed) or internally generating (imagined) words \cite{saeidi2021neural, livezey2021deep, wu2024speech, pescatore2025decoding}. Decoding covert speech remains harder, because the neural responses are weaker, noisier, and more difficult to collect in large amounts \cite{panachakel2021decoding, duraisamy2025transfer, maghsoudi2026zero}. One promising way around this limitation is to transfer knowledge between speech conditions. Previous work has shown that neural responses to vocalized, mimed, and imagined speech share common structure \cite{lee2019eeg}, and that decoders trained on one condition can be applied to another while still reaching reasonable performance \cite{maghsoudi2026relating}. However, while the similarity between these neural responses has been studied directly, little is known about how a decoder processes them internally. When the same decoder receives vocalized rather than mimed brain activity, which of its internal components change, and which of these changes actually affect the decoded speech? 

In this work, we leverage mechanistic interpretability to study how a brain-to-speech decoder handles neural responses from different speech conditions. Mechanistic interpretability provides a suite of tools for identifying which internal components of a trained network causally contribute to its predictions \cite{shimizu2025interpretable,mi5,mi3,mi2}. In this work, we rely on activation patching~\cite{activation_patching3}, a technique that replaces a network's internal activations on one input with those recorded on another input and measures the resulting change in its output. Specifically, we pass brain activity from a sentence produced in one condition (e.g., vocalized) through a decoder trained on another condition (e.g., mimed), and insert the activity of selected internal units (neurons) into the decoder's processing of the same sentence in its own condition. If decoding improves, those neurons carry information from the source condition that is useful for the target condition. We ask two questions: (1) Is transfer between conditions carried by a few individual neurons, or by groups of neurons acting together? and (2) Are the same neurons used across sentences and different source conditions, or does each case rely on different neurons? 

To answer these questions, we use the VocalMind dataset \cite{he2025vocalmind}, which contains stereotactic EEG (sEEG) recordings from a participant who produced the same sentences in vocalized, mimed, and imagined form. Because the sentence is identical across conditions, differences in the decoder's internal activity can be attributed to the speech condition.

\begin{figure*}[t]
  \centering
  \includegraphics[page=2, width=450pt]{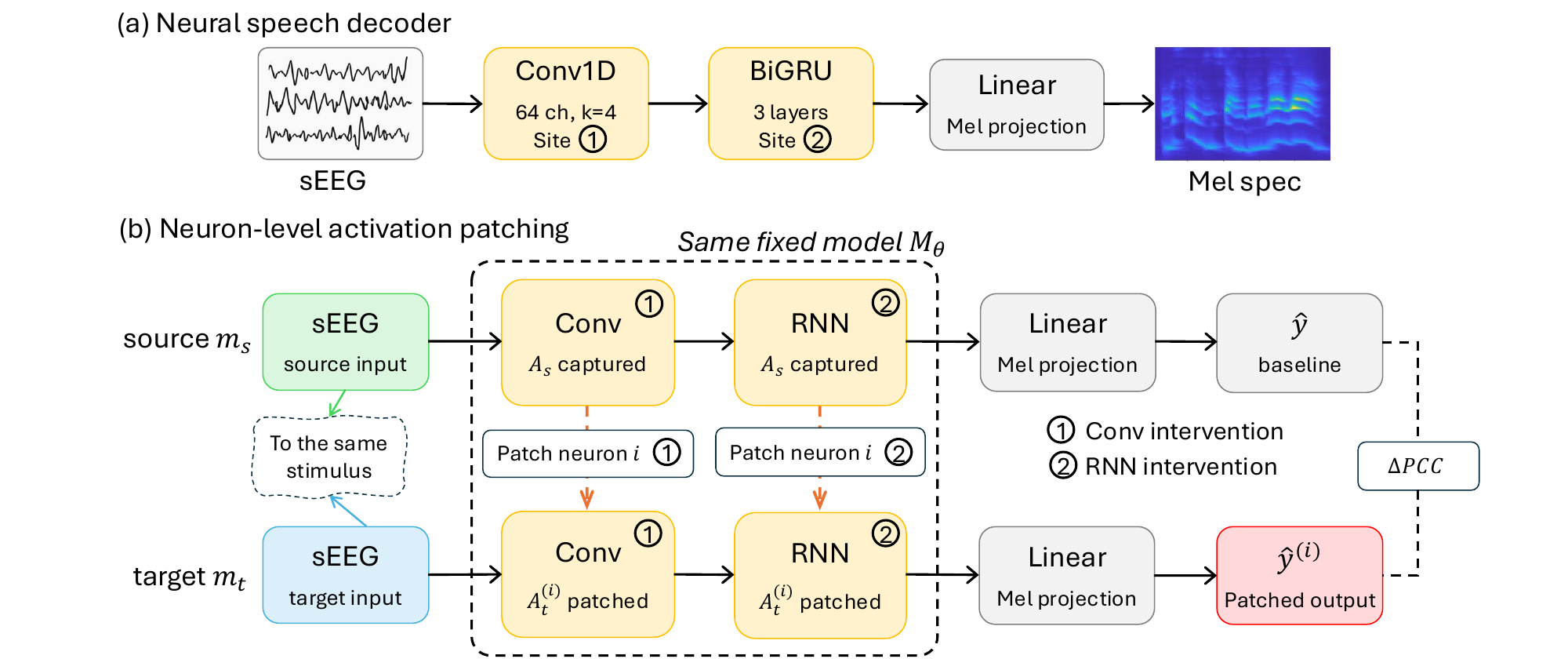}
  \caption{(a) Neural speech decoder with Conv1D and BiGRU intervention sites. (b) Neuron-level activation patching: source ($m_s$) and target ($m_t$) inputs from the same stimulus are passed through the target-mode model $M_\theta$. At a selected site, neuron $i$'s target activation is replaced with its source activation, and $\Delta\mathrm{PCC}$ measures the resulting change in performance.}
\label{fig:story}
\end{figure*}

To answer the first question, we measure the effect of inserting each neuron individually and then insert increasingly large groups of the most effective neurons. To answer the second, we identify the most effective neurons for each sentence and source condition and measure how often the same neurons reappear, compared with chance. We perform both analyses at an early stage of the decoder, which extracts short-term features, and a later stage, which integrates information over longer time spans.

We find that no single neuron is responsible for transfer between conditions. Instead, transfer arises from small groups of neurons. Brain activity from vocalized speech is the most useful source for decoding both mimed and imagined speech. The early stage relies on different neurons for each source condition, whereas the later stage reuses many of the same neurons across source conditions and sentences. Our findings pave the way for more data-efficient covert speech decoders, for example through training objectives that align covert representations with vocalized ones at the identified neurons, or by sharing the later stage across conditions while adapting only the early stage to each condition.

\section{Preliminaries and Approach}
\label{sec:contrib}
To investigate how a brain-to-speech decoder internally handles neural responses from different speech conditions, we use the three conditions as different inputs to the same decoder. Since each sentence is produced in every condition, the decoder's internal activity for one condition can be swapped into its processing of another, and the effect on the decoded speech can be measured. The sentences are identical; any change reflects differences between speech forms, so paired conditions act as input pairs for activation patching~\cite{activation_patching3}.


\textbf{Problem Setting.}
We consider three speech conditions (modes): vocalized, mimed, and imagined. For a source mode $m_s$ (patch-from) and target mode $m_t$ (patch-to), we consider paired sEEG inputs $(x^{(m_s)},x^{(m_t)})$ corresponding to the same sentence (Fig. \ref{fig:story}). Both are passed through the same decoder model $M_\theta$ trained on the target mode $m_t$. At layer $\ell$, the source and target activations are
$A_s=a_\ell(x^{(m_s)})$ and $A_t=a_\ell(x^{(m_t)}) \in
\mathbb{R}^{T_\ell\times D_\ell}$, where $T_\ell$ is the number of time steps
and $D_\ell$ the number of neurons (hidden units). The unpatched baseline prediction is $\hat{y}^{\mathrm{base}}=M_\theta(x^{(m_t)})$, and $y$ denotes the ground-truth mel-spectrogram of the sentence.

\begin{figure}[!b]
  \centering
  \includegraphics[width=\linewidth]{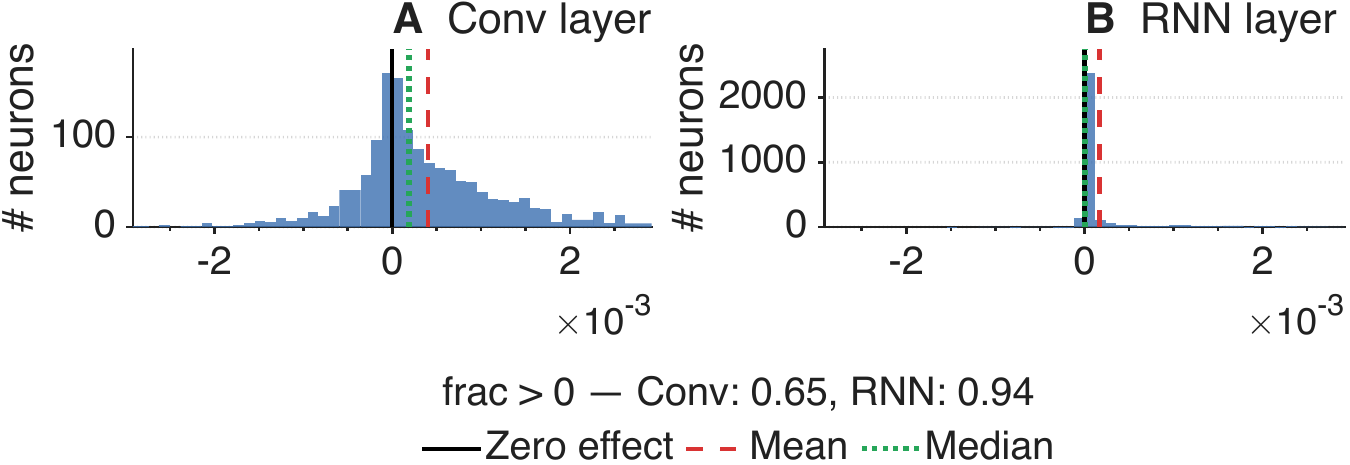}
  \caption{Distribution of single-neuron patching effects for
Vocalized$\rightarrow$Mimed, pooled across neurons and folds.
Left: convolutional layer; right: RNN layer. Effects concentrate near zero with broader distribution at the convolutional layer.}
  \label{fig:part1_main}
\end{figure}

\begin{figure*}[!t]
  \centering
  \includegraphics[width=420pt]{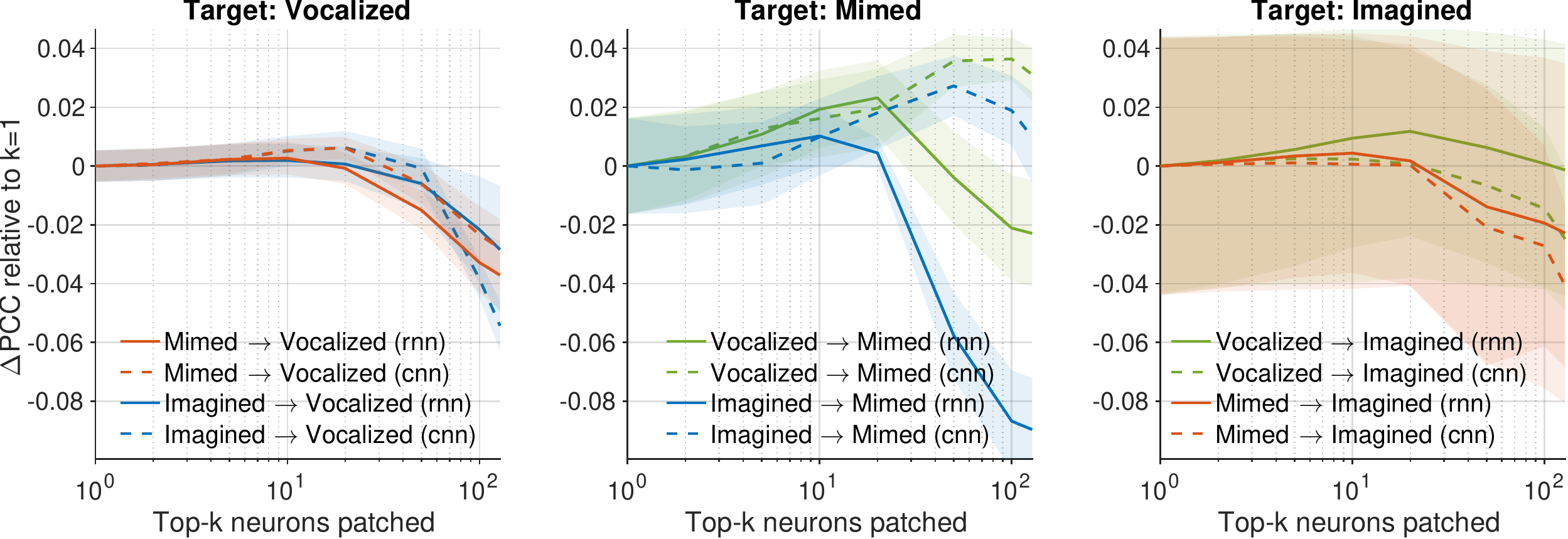}
  \caption{Top-$k$ neuron saturation curves across modality pairs and layers. $\Delta\mathrm{PCC}$ relative to $k{=}1$ is shown as a function of the number of patched neurons for target modalities. Solid: RNN; dashed: convolutional; shaded: $\pm$SEM across folds. For Mimed and Imagined targets, curves rise and then degrade, indicating transfer through compact neuron subsets. No improvement is observed for Vocalized targets, while patching from Vocalized activations yields the largest gains for covert targets.}
  \label{fig:part2_main}
\end{figure*}

\textbf{Activation Patching Framework.}
To measure how the activity of a set of neurons from another speech condition affects target-mode decoding, we replace their target activation trajectories with the corresponding source activations. For a neuron set $S \subseteq \{1,\ldots,D_\ell\}$:
\begin{equation}
\tilde{A}_t^{(S)}[t,j] =
\begin{cases}
A_s[t,j], & j\in S,\\
A_t[t,j], & \mathrm{otherwise},
\end{cases}
\quad \forall t.
\end{equation}
We continue the target-mode forward pass using $\tilde{A}_t^{(S)}$ to obtain the patched prediction $\hat{y}^{(S)}$, and compare it with the baseline:
\begin{equation}
\Delta\mathrm{PCC}(S)
=
\mathrm{PCC}(\hat y^{(S)},y)
-
\mathrm{PCC}(\hat y^{\mathrm{base}},y),
\end{equation}
where PCC is the Pearson correlation between predicted and ground-truth mel-spectrograms. Single-neuron patching is the special case $S=\{i\}$, written $\Delta\mathrm{PCC}_i$. Positive $\Delta\mathrm{PCC}$ indicates that the source activity of the neurons in $S$ improves target-mode decoding, measuring its ability to causally substitute for the target representation within the same decoder.

\textbf{Dataset.}
We use the VocalMind dataset \cite{he2025vocalmind}, containing sEEG recordings from one participant producing 100 Mandarin sentences, each repeated twice under vocalized, mimed, and imagined speech conditions. 

\textbf{Decoder.}
We use the neural speech decoder of \cite{he2025vocalmind}, consisting of a Conv1D encoder which extracts local spatiotemporal features from the raw sEEG signal, a three-layer bidirectional GRU~\cite{cho2014learning} with 256 hidden units per direction that integrates information over longer temporal contexts, and a linear layer predicting 80-dimensional mel-spectrograms, which are then converted to speech using HiFi-GAN~\cite{kong2020hifi}. We patch at two sites: the convolutional encoder output (\texttt{conv}, $D_\ell=64$), representing the early stage, and the recurrent network output (\texttt{rnn}, $D_\ell=512$), representing the later stage.

\textbf{Evaluation.} We use six-fold cross-validation over 200 samples per condition and perform all patching analyses on held-out test samples, for all source-target mode pairs. Results are aggregated across folds as mean and standard deviation of $\Delta\mathrm{PCC}$.


\textbf{Experiment 1: Individual Neurons vs. Neuron Groups.}
To test whether cross-condition transfer is carried by individual neurons, we patch each neuron separately ($S=\{i\}$) and examine the distribution of $\Delta\mathrm{PCC}_i$ across neurons and samples. If a single neuron drove transfer, its effect would separate clearly from the rest of the population.

To test whether transfer instead emerges from the collective effects of multiple neurons, we perform top-$k$ neuron patching. For each layer $\ell$ and source-target pair $(m_s\rightarrow m_t)$, we rank neurons by their mean single-neuron effect across samples, $\overline{\Delta\mathrm{PCC}}_i$, producing an ordering $\pi$, where $\pi(1)$ is the neuron with the largest effect. We define $S_k=\{\pi(1),\ldots,\pi(k)\}$ as the $k$ highest-ranked neurons, jointly patch them, and evaluate $\Delta\mathrm{PCC}(S_k)$ as a function of $k$. A rapid rise and plateau indicates that causal information is concentrated in a compact subset, while gradual improvement suggests a more distributed representation; degradation after a peak suggests that adding neurons introduces interference.

\textbf{Experiment 2: Reuse of Causal Neurons.}
To test whether the same neurons support transfer across different sentences, we define the \emph{winner neuron} for each input sample, for a given source-target pair and layer, as the neuron whose individual patching produces the largest $\Delta\mathrm{PCC}$. We measure reuse using the number of unique winners and top-$k$ coverage (the fraction of input samples accounted for by the $k$ most frequent winners). Fewer unique winners and higher coverage indicate greater neuron reuse across input samples.

To test whether different source conditions recruit the same neurons when targeting the same mode, we rank neurons by their winner count across input samples, breaking ties by mean winner $\Delta\mathrm{PCC}$. For a fixed target $m_t$ and two source modes $m_s^{(1)}$ and $m_s^{(2)}$, we define $S_k^{(1)}$ and $S_k^{(2)}$ as their $k$ highest-ranked neurons and compute $J(k)=|S_k^{(1)}\cap S_k^{(2)}|\,/\,|S_k^{(1)}\cup S_k^{(2)}|$. We compare overlap against a permutation null by randomly drawing two sets of $k$ neurons from the $D_\ell$ neurons at each layer 1000 times, and report the null mean and 95\% confidence interval as a function of $k$, accounting for differences in layer dimensionality.

\section{Results}

\textbf{No single neuron dominates cross-modal transfer.} Across all six source--target pairs and both layers, single-neuron patching
effects are small, with mean $\Delta\mathrm{PCC}$ on the order of $10^{-4}$.
Fig.~\ref{fig:part1_main} shows the representative
Vocalized $\rightarrow$ Mimed condition, where effects are concentrated near
zero and no individual neuron separates from the population. The convolutional
layer shows a broader distribution than the RNN, indicating greater variability
in single-neuron effects.

Despite their small magnitude, the direction of these effects varies across conditions. Vocalized$\rightarrow$Mimed shows the strongest positive bias at the RNN layer ($\mathrm{frac}{>}0=0.94$), whereas conditions targeting vocalized speech have fewer than half of neurons with positive effects. Thus, while the contribution of individual neurons depends on the transfer
direction, no single neuron accounts for cross-modal transfer. This motivates
testing whether transfer instead emerges from the collective effects of
multiple neurons.

\begin{figure}[t]
  \centering
  \includegraphics[width=\linewidth]{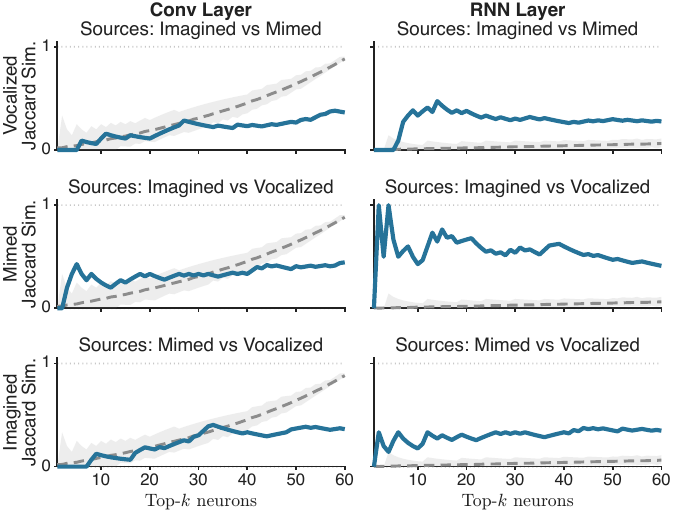}
  \caption{Top-$k$ neuron overlap across source modes. Jaccard similarity between top-$k$ causal neuron sets from two source modes for three target conditions (rows) and two layers (columns). Dashed line/shaded band: permutation-null mean and 95\% CI. Convolutional overlap falls at or below the null, whereas RNN overlap is far above the null across all targets, indicating a shared recurrent causal subspace\protect\footnotemark{}. Overlap is strongest for Mimed targets, reaching $\sim$0.8 at small $k$.}
  \label{fig:part4_main}
\end{figure}
\footnotetext{Here, causal subspace refers to activation directions that directly affect the model’s output.\cite{meloux2025everything}}

\textbf{cross-modal transfer emerges from compact neuron subsets.} Motivated by the absence of dominant single neurons, we jointly patch ranked neuron subsets (Fig.~\ref{fig:part2_main}). For Mimed and Imagined targets, performance initially improves and then degrades as $k$
increases, indicating that transfer is concentrated in compact neuron subsets and adding neurons beyond these subsets introduces interference. No improvement is observed for Vocalized targets.

Convolutional curves generally peak at larger $k$ than RNN curves for Mimed and Imagined targets, suggesting more distributed contributions at the convolutional layer and more compact subsets at the recurrent layer.
Imagined$\rightarrow$Mimed yields modest gains, whereas Mimed$\rightarrow$Imagined produces little improvement.

\textbf{Vocalized speech is the most useful source.} Patching Vocalized activations into Mimed or Imagined targets produces the largest gains, whereas Mimed or Imagined sources yield smaller or negligible improvements. This asymmetry may reflect the richer acoustic information available in neural responses during vocalized speech, where participants receive auditory input from their own speech \cite{hickok2007cortical, houde2015cortical}.

\textbf{Recurrent neurons are reused across samples.}  We next ask whether these neurons are reused (1) across input samples and (2) across source modes. Sample-level winner analysis addresses the first question and shows a layer-wise pattern: across source--target pairs, influential RNN neurons are reused more consistently across samples than convolutional neurons ($p<0.05$). The strongest reuse occurs for Vocalized$\rightarrow$Mimed at the RNN layer, with the fewest unique winners (69 across 200 samples) and the top five accounting for 35\% of samples. Together with the strong cross-source overlap for Mimed targets, this suggests that mimed speech shares representational structure with vocalized and imagined speech, consistent with its proposed intermediate position: it retains articulatory structure of vocalized speech while lacking overt acoustic output, like imagined speech \cite{maghsoudi2026relating, martin2018decoding}.

\textbf{Recurrent neurons are shared across source modalities.} Fig.~\ref{fig:part4_main} answers the second question by comparing the top-$k$ neuron sets obtained from two different source modes for the same target. At the convolutional layer, Jaccard overlap remains at or below the permutation null across all targets (Fig.~\ref{fig:part4_main}, left), indicating largely source-specific neuron recruitment. In contrast, RNN overlap is far above the null across all targets (roughly $5$--$8\times$ chance, stabilizing around $0.3$--$0.4$), indicating that source modes partially converge on the same recurrent neurons (Fig.~\ref{fig:part4_main}, right). Mimed targets show the strongest RNN overlap, reaching $\sim$0.8 at small $k$ before stabilizing around $0.4$--$0.6$.


\section{Discussion and Limitations}
Our results show that cross-modal transfer between speech conditions is carried by small groups of neurons rather than single units. Convolutional representations are more source- and sample-specific, whereas recurrent representations contain a more compact shared causal subspace that is reused across source modes and input samples.

The intermediate position of mimed speech may explain both the strong Vocalized$\rightarrow$Mimed transfer and the high overlap between source modes when mimed speech is the target.

These findings suggest that shared representations, especially those learned from vocalized speech, could support data-efficient covert speech decoding through training objectives that encourage cross-modal transfer.

Our study is limited to one participant and one Conv1D--BiGRU architecture, providing a controlled setting for isolating cross-modal representations. Future work should test whether these mechanisms generalize across participants and language-model-based decoders, where linguistic priors may interact with neural evidence, and relate the identified causal subspaces to the anatomical organization of electrodes.

\vfill\pagebreak

{\small
\bibliographystyle{IEEEbib}
\bibliography{refs}
}


\end{document}